\documentclass{article}
\usepackage{spconf,amsmath,graphicx}
\usepackage{amssymb,bm}
\usepackage{url}
\newcommand{\mat}[1]{\boldsymbol{#1}} % matrix

\title{Look globally, age locally: Face aging with an attention mechanism}
\name{Haiping Zhu \qquad Zhizhong Huang \qquad Hongming Shan \qquad Junping Zhang}

\address{Shanghai Key Lab of Intelligent Information Processing, \\ School of Computer Science, Fudan University, China, 200433 \\
\{hpzhu14, zzhuang19, hmshan, jpzhang\}@fudan.edu.cn
}

\begin{document}
%\ninept
%
\maketitle
\begin{abstract}
Face aging is of great importance for cross-age recognition and entertainment-related applications. Recently, conditional generative adversarial networks (cGANs) have achieved impressive results for face aging. Existing cGANs-based methods usually require a pixel-wise loss to keep the identity and background consistent. However, minimizing the pixel-wise loss between the input and synthesized images likely resulting in a ghosted or blurry face. To address this deficiency, this paper introduces an Attention Conditional GANs (AcGANs) approach for face aging, which utilizes attention mechanism to \emph{only} alert the regions relevant to face aging. In doing so, the synthesized face can well preserve the background information and personal identity without using the pixel-wise loss, and the ghost artifacts and blurriness can be significantly reduced. Based on the benchmarked dataset Morph, both qualitative and quantitative experiment results demonstrate superior performance over existing algorithms in terms of image quality, personal identity, and age accuracy.
\end{abstract}
\begin{keywords}
Face Aging; Attention Mechanism; Conditional GANs; Pixel-wise Loss; Adversarial Training.
\end{keywords}
\section{Introduction}
Face aging, also known as age progression, aims to render a given face image with natural aging effects under a certain age or age group. In recent years, face aging has attracted major attention due to its extensive use in numerous applications, entertainment~\cite{fu2010age}, finding missing children~\cite{kemelmacher2014illumination}, cross-age face recognition~\cite{park2010age}, etc. Although impressive results have been achieved recently~\cite{zhang2017age,
wang2018face,liu2019attribute,palsson2018generative,yang2018learning}, there are still many challenges due to the intrinsic complexity of aging in nature and the insufficient labeled aging data. Intuitively, the generated face images should be photo-realistic, \emph{e.g.}, without serious ghosting artifacts. In addition to that, the face aging accuracy and personal identify permanence of the generated face images should be guaranteed simultaneously.

Recently, the generative adversarial networks (GANs)~\cite{goodfellow2014generative} have shown an impressive ability in generating synthetic images~\cite{gauthier2014conditional} and face aging~\cite{wang2016recurrent,zhang2017age,wang2018face,liu2019attribute,palsson2018generative,yang2018learning}. These approaches render faces with more natural aging effects in terms of high quality, identity consistency, and aging accuracy compared to the previous conventional solutions, such as prototype-based~\cite{kemelmacher2014illumination} and physical model-based methods~\cite{suo2012concatenational,suo2009compositional}. However, the problems have not been completely solved. For example, Zhang~\emph{et al.}~\cite{zhang2017age} first proposed a conditional adversarial autoencoder (CAAE) for face aging by traversing on the face manifold in low dimension, but it cannot keep the identity information of generated faces well. To solve this problem, Yang~\emph{et al.}~\cite{yang2018learning} and Wang~\emph{et al.}~\cite{wang2018face} proposed a condition GANs with a pre-trained neural network to preserve the identity of generated images. 
Most existing GANs-based methods usually train the model with the pixel-wise loss~\cite{zhang2017age,yang2018learning} to preserve identity consistency and keep background information. But to minimize the Euclidean distance between the synthesized images and the input images will easily cause the synthesized images becoming ghosted or blurred~\cite{isola2017image}. In particular, this problem would be more severe if the gap between the input age and the target age becomes larger. 

Inspired by the success of attention mechanism in image-to-image translation~\cite{pumarola2018ganimation}, in this paper, we propose an {\bf A}ttention {\bf C}onditional {\bf GANs} (AcGANs) to tackle these issues mentioned-above. Specifically, the proposed AcGANs consists of a generator $G$ and a discriminator $D$. The generator $G$ receives an input image and a target age code and the output of the generator contains an attention mask and a color mask. The attention mask learns the modified regions relevant to face aging and the color mask learns how to modify. The final output of the generator is a combination of the attention and the color masks. Since the attention mechanism only modifies the regions relevant to face aging, it can preserve the background information and the personal identity well without using the pixel-wise loss in training. The discriminator $D$ consists of an image discriminator and an age classifier, aiming to make the generated face be more photo-realistic and guarantee the synthesized face lies in the target age group.

The main contributions of this paper are: i) We propose a novel approach for face aging, which utilizes an attention mechanism to modify the regions relevant to face aging. Thus,
the synthesized face can well preserve the background information and personal identity without using pixel-wise loss, and the ghost artifacts and blurriness can be significantly reduced. ii) Both qualitative and quantitative experiment results on Morph demonstrate the effectiveness of our model in terms of image quality, personal identity, and age accuracy.

\section{The Proposed Methods}
We divide faces with different ages into 5 nonoverlapping groups, \emph{i.e.,} 11-20, 21-30, 31-40, 41-50, and 50+. Given a face image $\mat{x}\in\mathbb{R}^{h\times w\times 3}$, where $h$ and $w$ are the hight and width of the image, respectively. We use a one-hot label $\mat{y}\in \mathbb{R}^{1\times 5}$ to indicate the age group that $\mat{x}$ belongs to. The aim is to learn a generator $G$ to generate a synthesized face image $\mat{x}_{t}$ that lies in target age group $\mat{y}_t$, looks realistic, and has the same identity as the input face image $\mat{x}_s$.

\subsection{Network Architecture}
\begin{figure}[!ht]
\centering
\includegraphics[width=0.98\linewidth, height=0.42\linewidth, clip=true, trim=50 90 40 60]{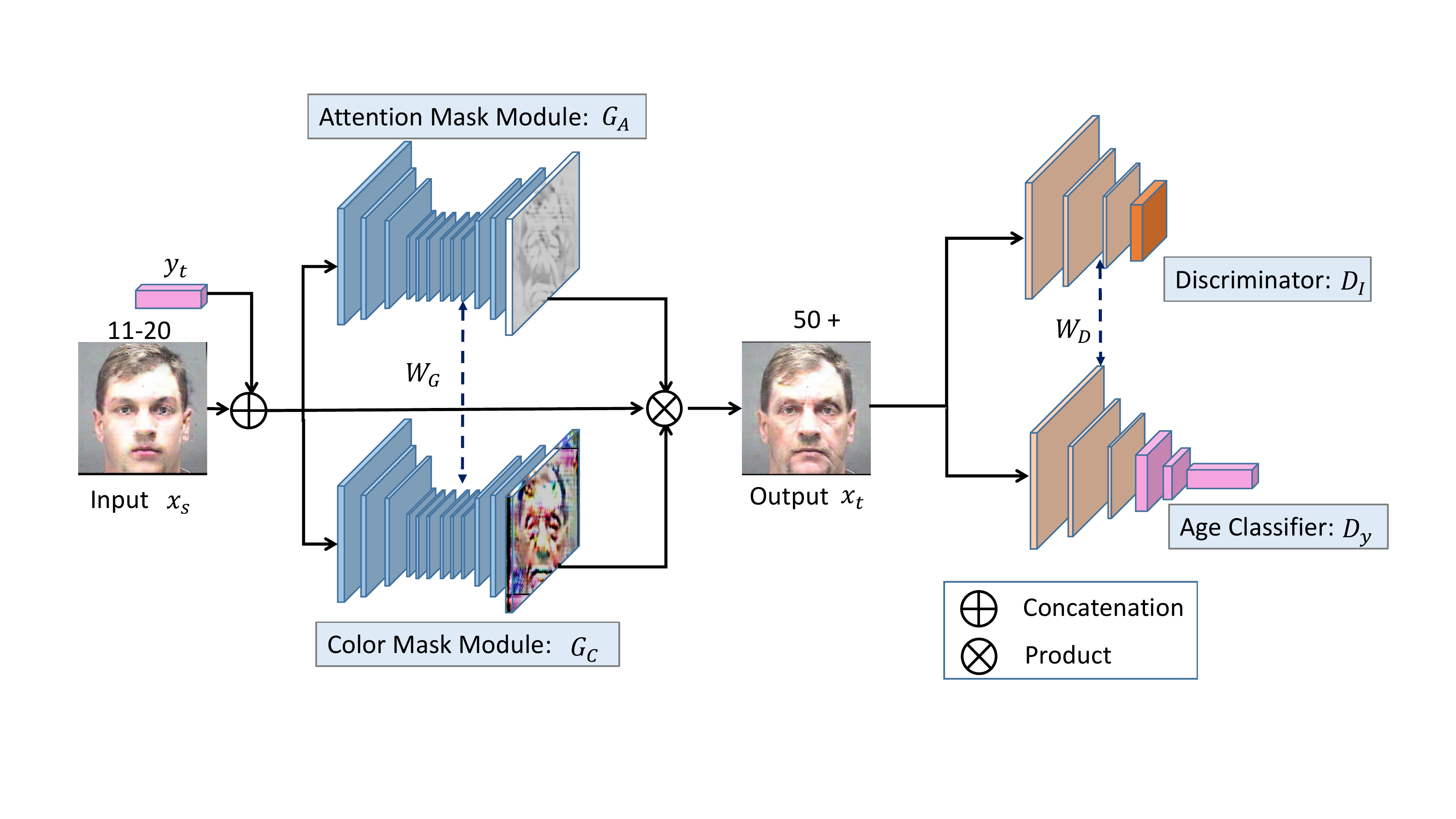}
\caption{The architecture of the proposed method. In our network, color mask module $G_C$ and attention mask module $G_A$ share parameters $\mat{W}_G$, except for the last layer. Similarly, discriminator $D_I$ and age classifier $D_y$ share parameters $\mat{W}_D$.}
\label{network}
\end{figure}

The proposed approach, shown in Fig.~\ref{network}, consists of two main modules: i) A generator $G$ is trained to generate a synthesized face $\mat{x}_t$ with target age $\mat{y}_t$; ii) A discriminator $D$ aims to make $\mat{x}_t$ looks realistic and guarantee $\mat{x}_t$ lie in target age group $\mat{y}_t$.

\noindent\textbf{Generator } Given a input face image $\mat{x}_s \in\mathbb{R}^{h\times w\times 3}$ and a target age $\mat{y}_t\in\mathbb{R}^{1\times 5}$, we need to pad the one-hot label $\mat{y}_t$ into $\mathbb{R}^{h\times w\times 5}$. Then, we form the input of generator as a concatenation $(\mat{x}_s, \mat{y}_t)\in\mathbb{R}^{h\times w\times (3 + 5)}$. 

One key ingredient of our approach is to make $G$ focus on those regions of the image that are relevant to face aging and keep the background information unchanged and preserve identity consistency. For this purpose, we have embedded an attention mechanism to the generator. Concretely, instead of regressing a full image, our generator outputs two masks, an attention mask $\mat{A}$ and a color mask $\mat{C}$. The final generated image can be obtained as:
\begin{equation}
\mat{x}_{t} = G(\mat{x}_s,\mat{y}_t) = (\mat{I} - \mat{A})\otimes \mat{C} + \mat{A}\otimes\mat{x}_s,
\end{equation}
where $\otimes$ denotes the element-wise product, $\mat{A} = G_A(\mat{x}_{s},\mat{y}_t)\in\{0,\dots,1\}^{h\times w\times 1}$, and $\mat{C} = G_C(\mat{x}_s,\mat{y}_t)\in \mathbb{R}^{h\times w\times 3}$. The mask $\mat{A}$ indicates to which extend each pixel of the $\mat{C}$ contributions to the generative image $\mat{x}_t$.

\noindent\textbf{Discriminator } This module consists of an image discriminator $D_I$ and an age classifier $D_y$, aiming to make the generated face be realistic and guarantee the synthesized face lies in the target age group. Note that $D_I$ and $D_y$ share parameters $\mat{W}_D$, as shown in Fig.~\ref{network}, which makes the performance of the discriminator $D$ improve significantly.

\subsection{Loss Function}
The defined loss function includes three terms: 1) \textbf{The adversarial loss} proposed by Gulrajani \emph{et al.}~\cite{gulrajani2017improved} that pushed the distribution of the generated images to the distribution of the training images; 2) \textbf{The attention loss} to drive the attention masks to be smooth and prevent them from saturating; 3) \textbf{The age classification loss} to make the generated facial image more accurate in age classification.

\noindent\textbf{Adversarial Loss } To learn the parameters of the generator $\mat{G}$, we utilize the modification of the standard GAN algorithm~\cite{goodfellow2014generative} proposed by Wasserstein GAN with gradient penalty (GAN-GP)~\cite{gulrajani2017improved}. Specifically, the original GAN formulation is based on the Jenson-Shannon (JS) divergence loss function and aims to maximize the probability of correctly classifying real and fake images while the generator tries to fool the discriminator. This loss is potentially not continuous for the parameters of the generator and can locally saturate leading to vanishing gradients in the discriminator. This is addressed in WGAN~\cite{arjovsky2017wasserstein} by replacing JS with the continuous Earth Mover Distance. To maintain a Lipschitz constraint, WGAN-GP~\cite{gulrajani2017improved} added a gradient penalty for the critic network computed as the norm of the gradients for the critic input.

Formally, let $\mathbb{P}_{x_s}$ be the distribution of the input image $\mat{x}_s$, and $\mathbb{P}_{\widetilde{\mat{x}}}$ be the random interpolation distribution between $\mathbb{P}_{x_s}$ and $\mathbb{P}_{x_t}$. Then, the adversarial loss $\mathcal{L}_{adv}(G, D_I, \mat{x}_s, \mat{y}_s)$ can be written as:
\begin{align}
\mathcal{L}_{adv} = 
&\mathbb{E}_{\mat{x}_s\sim\mathbb{P}_{x_s}}[D_I(\mat{x}_s)] - \mathbb{E}_{\mat{x}_s\sim\mathbb{P}_{x_s}}[D_I(G(\mat{x}_s,\mat{y}_t))] \notag \\
& - \lambda_{gp}\mathbb{E}_{\widetilde{\mat{x}}\sim\mathbb{P}_{\widetilde{x}}}\Big[(\lVert\nabla_{\widetilde{\mat{x}}}D_I(\widetilde{\mat{x}})\rVert - 1)^2\Big],
\end{align}
where $\lambda_{gp}$ is a penalty coefficient.

\noindent\textbf{Attention Loss } Note that when training the model, we do not have ground-truth annotation for the attention masks $\mat{A}$. Similarly as for the color masks $\mat{C}$, they are learned from the resulting gradients of the discriminative module and the age classification loss. However, the attention masks can easily saturate to 1, which makes that the attention module does not effect. To prevent this situation, we regularize the mask with a $l_2$-weight penalty. Besides, to enforce smooth spatial color transformation when combining the pixel from the input image and the color transformation $\mat{C}$, we perform a \emph{Total Variation Regularization} over $\mat{A}$. The attention loss $\mathcal{L}_{att}(G, \mat{x}_s, \mat{y}_t)$ can be defined as:
{\scriptsize
\begin{align}
\mathcal{L}_{att} = &\lambda_{TV}\mathbb{E}_{\mat{x}_s\sim\mathbb{P}_{x_s}}\Bigg[\sum_{i,j}^{h,w}[(\mat{A}_{i+1,j} - \mat{A}_{i,j})^2 + (\mat{A}_{i,j+1} - \mat{A}_{i,j})^2]\Bigg] \notag \\
& + \mathbb{E}_{\mat{x}_t\sim\mathbb{P}_{x_t}}\Big[\lVert\mat{A}\rVert_2\Big],
\end{align}}
where $\mat{A} = G_A(\mat{x},\mat{y}_t)$ and $\mat{A}_{i,j}$ is the $i,j$ entry of $\mat{A}$. Besides, $\lambda_{TV}$ is a penalty coefficient.

\noindent\textbf{Age Classification Loss } While reducing the image adversarial loss, the generator must also reduce the age error by the age classifier $D_y$. The age classification loss is defined with two components: an age estimation loss with fake images used to optimize G, and an age estimation loss of real images used to learn the age classifier $D_y$. This loss $\mathcal{L}_{cls}(G, D_y, \mat{x}_s, \mat{y}_t, \mat{y}_s)$ is computed as:
{\scriptsize
\begin{equation}
\mathcal{L}_{cls} = \mathbb{E}_{\mat{x}_s\sim\mathbb{P}_{x_s}}\Big[\ell(D_y(G(\mat{x}_s,\mat{y}_t)), \mat{y}_t) + \ell(D_y(\mat{x}_s), \mat{y}_s)\Big],
\end{equation}}
where $\mat{y}_s$ is the label of input image $\mat{x}_s$, $\ell(\cdot)$ corresponds to a softmax loss.

\noindent\textbf{Final Loss } To generate the target age image $\mat{x}_t$, we build a loss function $\mathcal{L}$ by linearly combining all previous losses:
\begin{align}
\mathcal{L} = &\lambda_{adv}\mathcal{L}_{adv}(G, D_I, \mat{x}_s, \mat{y}_t) + \lambda_{att}\mathcal{L}_{att}(G, \mat{x}_s, \mat{y}_t) \notag \\ 
&+ \lambda_{cls}\mathcal{L}_{cls}(G, D_y, \mat{x}_s, \mat{y}_t, \mat{y}_s),
\end{align}
where $\lambda_{adv}$, $\lambda_{att}$ and $\lambda_{cls}$ are the hyper-parameters that control the relative importance of every loss term. Finally, we can define the following minimax problem:
\begin{equation}
G^{*} = \arg\min_{G}\max_{D\in\mathcal{D}} \mathcal{L},
\end{equation}
where $G^*$ draws samples from the data distribution. Additionally, we constrain our discriminator $D$ to lie in $\mathcal{D}$, which represents the set of 1-Lipschitz functions.

\section{Experiments}
In this section, we introduce our implementation details and then evaluate our proposed model both qualitatively and quantitatively on a large public dataset Morph~\cite{ricanek2006morph}, which contains 55,000 face images of 13,617 subjects from 16 to 77 years old. To better demonstrate the superiority in preserving identity features of our methods, we have also compared the two state-of-the-art methods: Conditional Adversarial Autoencoder Network~(CAAE)~\cite{zhang2017age} and Identity-Preserved Conditional Generative Adversarial Networks~(IPCGANs)~\cite{wang2018face}.

\begin{figure}[ht!]
\centering
\includegraphics[width=1\linewidth, height=0.5\linewidth, clip=true, trim=20 60 20 0]{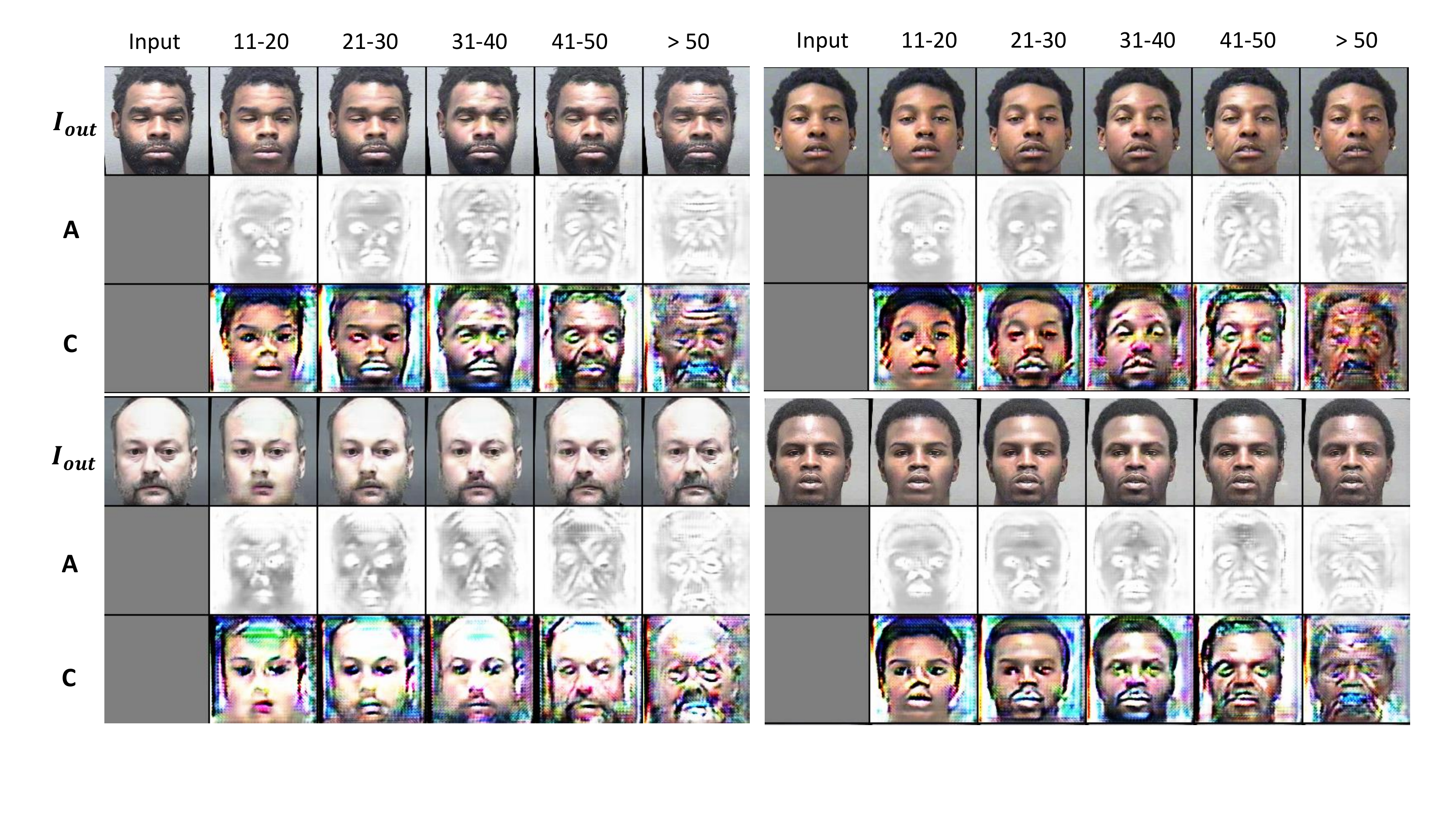}
\caption{Illustration of generation results by the proposed AcGANs method. For each subject, the first row images are the generative face aging images, the second row is the details of the intermediate attention mask \textbf{A}, and the third row is the color mask \textbf{C}.}
\label{attention_result}
\end{figure}

\subsection{Implementation Details}
Following prior works~\cite{zhang2017age, wang2018face, liu2019attribute}, before fed into the networks, the faces are (1) aligned by the ﬁve facial landmarks detected by MTCNN~\cite{zhang2016joint}, (2) cropped to $224 \times 224$ pixels of 10\% more area, thus not only hair but also beard are all covered, (3) divided into five age groups, \emph{i.e.}, 10-20, 21-30, 31-40, 41-50, 51+. Consequently, a total of 54,539 faces are collected and then we split Morph dataset into two parts, 90\% for training and the rest for testing without overlapping. The realization of AcGANs is based on the open-source ``PyTorch'' framework\footnote{The code has been released in https://github.com/JensonZhu14/AcGAN.}.

During training, we adopt an architecture similar with~\cite{liu2019attribute} which is shown in Fig. \ref{network}. Different from \cite{liu2019attribute}, our generator receives $224 \times 224 \times 3$ images and $224 \times 224 \times 5$ condition feature maps concatenated together along channel as input, which is larger than $128 \times 128$ of CAAE and IPCGANs, thus a more clear result is generated. Furthermore, the conditional feature maps are similar to one-hot code in some ways where only one of which is ﬁlled with ones while the rest are all ﬁlled with zeros. For IPCGANs, we first train the age classiﬁer which is ﬁnetuned based on AlexNet on the CACD~\cite{chen2014cross} and other parameters are set according to~\cite{zhang2017age}. For CAAE, we remove the gender information and use 5 age groups instead of for fair comparison. For AcGANs, we set $\lambda_{adv}$ to 10, while $\lambda_{att}$ is 2, $\lambda_{cls}$ is 100, $\lambda_{gp}$ is 10, and $\lambda_{TV}$ is $5e-5$, respectively. For all of them including AcGANs, we choose Adam to optimize both $G$ and $D$ with learning rate and batch-size set to $1e−4$ and 64, respectively. Thus we train the $G$ and $D$ in turn every iteration with total 100 epochs on four 2080 Ti GPU.

\subsection{Results on Morph}
In this subsection, we first visualize the aging process from the perspective of what AcGANs have learned from the input image, \emph{i.e.}, attention mask and color mask. As shown in Fig.~\ref{attention_result}, we select four face images from the test dataset randomly regardless of their original age group and exhibit the aging results in the first row while the second row is attention mask and the third row is color mask correspondingly. According to the attention mask, we can draw a convincing conclusion that AcGANs indeed learns which parts of the face should be aged. 

\begin{figure}[h!]
\centering
\includegraphics[width=0.92\linewidth, height=1.2\linewidth, clip=true, trim=0 0 0 0]{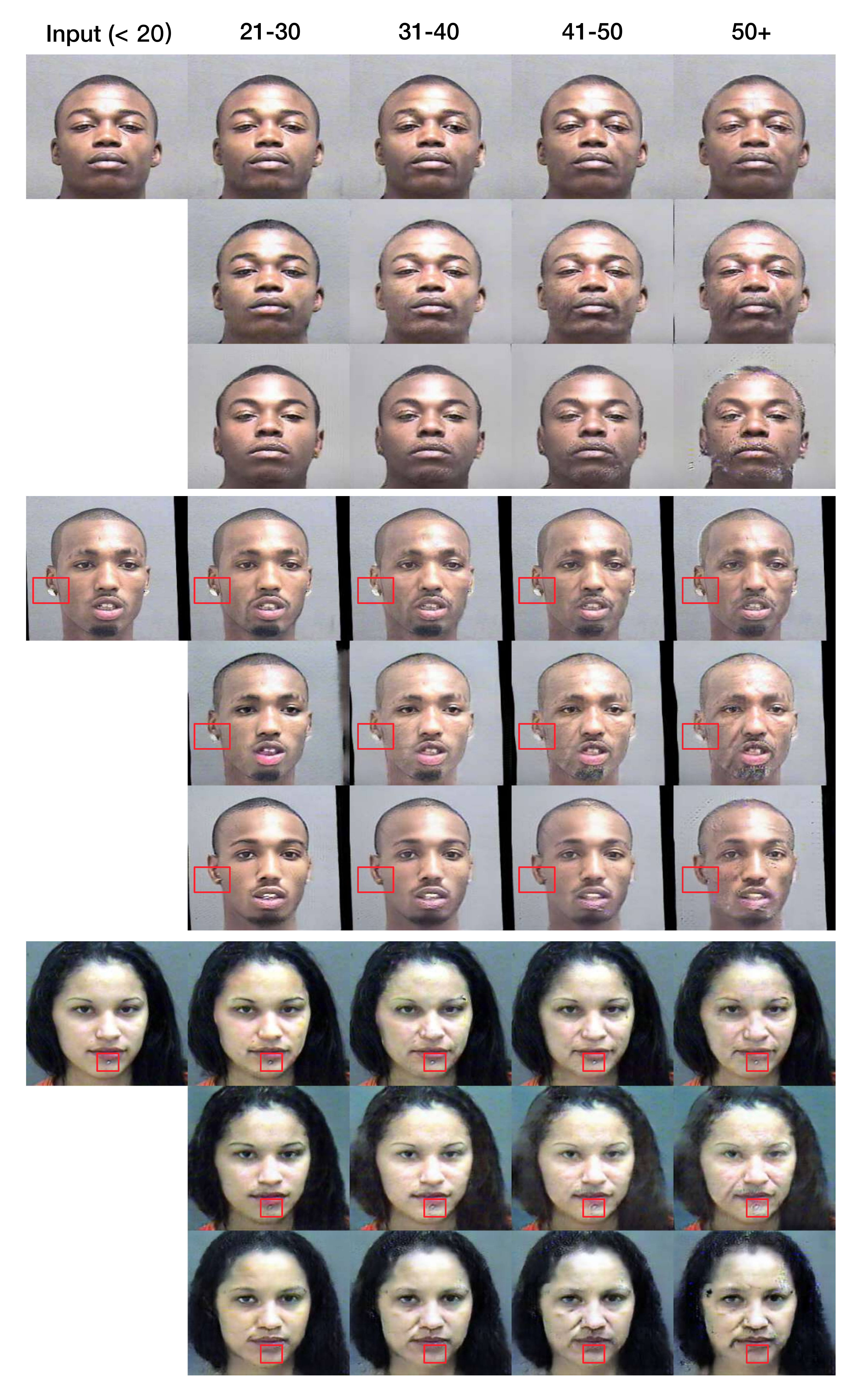}
\caption{Some synthesized faces generated by different methods. For each sample, from top to bottom, they are images generated by AcGANs, IPCGANs, and CAAE. The input age lines in [11-20] age group and the numbers above the images are the corresponding target age.}
\label{morph_result}
\end{figure}

We further qualitatively compare the generated faces of different methods in Fig.~\ref{morph_result}. All of the three generated results show that AcGANs has a more powerful capability of removing ghosted artifacts. Meanwhile, the adornments marked in the red rectangle of the last two faces are preserved integrally by AcGANs, which has proved that AcGANs has learned what should be aged in the face once again.

\subsection{Quantitative Comparison}
To avoid the suspicion that the limited images demonstrated in the paper, we have also evaluated all results quantitatively. In literature~\cite{wang2018face, liu2019attribute}, there are two critical evaluation metrics in age progression, \emph{i.e.} identity permanence and aging accuracy. We first generate the elder faces from young faces, \emph{i.e.} faces of 10-20 age group, and then evaluate them separately.

To estimate the aging accuracy, we use Face++ API~\cite{faceplusplus.com} to estimate the age distributions, \emph{i.e.}, mean value of both generic and generated faces in each age group, where less discrepancy between real and fake images indicates more accurate simulation of aging effects. For simplicity, we report the mean value of age distributions while the discrepancy with generic age distribution is shown in brackets (seen in Table~\ref{age_distribution}). For identity permanence, Face veriﬁcation experiments are also conducted on Face++ API, where high verification confidence and verification rate indicates a powerful performance to preserve identity information. From Table~\ref{face_verification} it can be seen that the top is verification confidence between ground truth young faces and their aging elder ones generated by AcGANs, and the bottom is verification rate between them which means the accuracy that they are the same person. The best values for each column of both Table~\ref{age_distribution} and Table~\ref{face_verification} are indicated in bold.

On Morph, it could be easily seen that our AcGANs consistently outperform CAAE and IPCGANs in two metrics during all four aging processes. Although IPCGANs has a better capability in preserving identity information, it generates worse inferior aging faces than CAAE, while CAAE fails to keep the original identity. However, AcGANs could not only achieve a better aging result but also preserve identity consistently in an advantageous position.
\begin{table}[h]
\centering
    \footnotesize
    \begin{tabular}{lrrrr}
    \hline
    \multicolumn{5}{c}{Estimated Age Distributions}                                                                                                                           \\ \hline
    Age group & \multicolumn{1}{c}{21-30}             & \multicolumn{1}{c}{31-40}             & \multicolumn{1}{c}{41-50}             & \multicolumn{1}{c}{50+}               \\ \hline
    Generic   & \multicolumn{1}{c}{25.12}             & \multicolumn{1}{c}{35.43}             & \multicolumn{1}{c}{44.72}             & \multicolumn{1}{c}{54.88}             \\
    CAAE~\cite{zhang2017age}   & 24.31(0.81)  & 31.02(4.41)  & 39.03(5.69)  & 47.84(7.04)  \\
    IPCGANs~\cite{wang2018face} & 22.38(2.74) & 27.53(7.90) & 36.41(8.31)  & 46.42(8.46)  \\
    AcGANs & \multicolumn{1}{c}{\textbf{25.92(0.80)}} & \multicolumn{1}{c}{\textbf{36.49(1.06)}} & \multicolumn{1}{c}{\textbf{40.59(4.13)}} & \multicolumn{1}{c}{\textbf{47.88(7.00)}} \\ \hline
    \end{tabular}
    \caption{Estimated Age Distributions (in years) on MORPH. Generic means that the mean value of each group is computed in the ground truth, while the number in brackets indicates the differences from generic mean age.}
    \label{age_distribution}
\end{table}

\begin{table}[h]
    \centering
        \small 
        \begin{tabular}{ccccc}
        \hline
        \multicolumn{1}{l}{} & 21-30                      & 31-40                      & 41-50                     & 50+                       \\ \cline{2-5} 
        \multicolumn{1}{l}{} & \multicolumn{4}{c}{Veriﬁcation Conﬁdence}                                                                       \\ \cline{2-5} 
        10-20                & 95.36                      & 94.78                      & 94.74                     & 93.44                     \\
        21-30                & -                          & 95.37                      & 95.28                     & 94.11                     \\
        31-40                & -                          & -                          & 95.65                     & 94.72                     \\
        41-50                & -                          & -                          & -                         & 95.26                     \\ \cline{2-5} 
        \multicolumn{1}{l}{} & \multicolumn{4}{c}{\begin{tabular}[c]{@{}c@{}}Veriﬁcation Rate (Threshold = 73.975,\\ FAR = 1e-5)\end{tabular}} \\ \cline{2-5} 
        CAAE~\cite{zhang2017age}      & 99.38                      & 97.82                      & 92.72                     & 80.56                     \\
        IPCGANs~\cite{wang2018face}     & \textbf{100}            & \textbf{100}            & \textbf{100}           & \textbf{100}           \\
        AcGANs                & \textbf{100}            & \textbf{100}            & \textbf{100}           & \textbf{100}           \\ \hline
        \end{tabular}
    \caption{Face verification results on Morph. The top is the verification confidence by AcGANs and the bottom is the verification rate for all methods. Noted that the generated and the input faces are considered as the same identity if the verification confidence is above the pre-defined threshold.}
    \label{face_verification}
\end{table}

\section{Conclusions}
In this paper, we propose a novel approach based on an attention mechanism for face aging. Since the attention mechanism only modifies the regions relevant to face aging, the proposed approach can well preserve the background information and the personal identity without using the pixel-wise loss, significantly reducing the ghost artifacts and blurring. Besides, the proposed approach is simple for it consists of only a generator and a discriminator sub-networks and can be learned without additional pre-trained models. Moreover, both qualitative and quantitative experiments validate the effectiveness of our approach.

% References should be produced using the bibtex program from suitable
% BiBTeX files (here: strings, refs, manuals). The IEEEbib.bst bibliography
% style file from IEEE produces unsorted bibliography list.
% -------------------------------------------------------------------------
% \bibliographystyle{IEEEbib}

\end{document}